\documentclass[sn-basic,iicol]{sn-jnl}

\usepackage{amsmath, amssymb, mathtools, bm}

\jyear{2023}%

\theoremstyle{thmstyleone}%

\theoremstyle{thmstyletwo}%

\theoremstyle{thmstylethree}%

\begin{document}

\title[Net]{Gaussian Mixture based Evidential Learning for Stereo Matching}

\author[1]{\fnm{Weide} \sur{Liu}}\email{weide.liu@childrens.harvard.edu}
\author[2]{\fnm{Xingxing} \sur{Wang}}\email{wangxingxing@outlook.com}
\author*[3]{\fnm{Lu} \sur{Wang}}\email{ wang\_lu@i2r.a-star.edu.sg}
\author[3]{\fnm{Jun} \sur{Cheng}}\email{cheng\_jun@i2r.a-star.edu.sg}
\author[3]{\fnm{Fayao} \sur{Liu}}\email{liu\_fayao@i2r.a-star.edu.sg}
\author[3]{\fnm{Xulei} \sur{Yang}}\email{yang\_xulei@i2r.a-star.edu.sg}

\affil[1]{\orgdiv{Boston Children's Hospital}, \orgname{Harvard Medical School}, \orgaddress{\city{MA}, \postcode{02115}, \country{USA}}}

\affil[2]{\orgname{Nanyang Technological University}, \orgaddress{\city{Singapore}, \postcode{639798}, \country{Singapore}}}

\affil[3]{\orgname{A*STAR}, \orgaddress{\city{Singapore}, \postcode{138634}, \country{Singapore}}}




\abstract{In this paper, we introduce a novel Gaussian mixture based evidential learning solution for robust stereo matching. 
Diverging from previous evidential deep learning approaches that rely on a single Gaussian distribution, our framework posits that individual image data adheres to a mixture-of-Gaussian distribution in stereo matching. This assumption yields more precise pixel-level predictions and more accurately mirrors the real-world image distribution. By further employing the inverse-Gamma distribution as an intermediary prior for each mixture component, our probabilistic model achieves improved depth estimation compared to its counterpart with the single Gaussian and effectively captures the model uncertainty, which enables a strong cross-domain generation ability.
We evaluated our method for stereo matching by training the model using the Scene Flow dataset and testing it on KITTI 2015 and Middlebury 2014.
The experiment results consistently show that our method brings improvements over the baseline methods in a trustworthy manner. Notably, our approach achieved new state-of-the-art results on both the in-domain validated data and the cross-domain datasets, demonstrating its effectiveness and robustness in stereo matching tasks. The code in this paper will be released on GitHub.}


\maketitle

\section{Introduction}
Deep learning-based neural networks are increasingly employed in safety-critical domains such as self-driving cars and robotics. The accurate assessment of model uncertainty is essential for their broader acceptance. Providing precise and well-calibrated uncertainty estimation is crucial for assessing confidence levels, detecting domain shifts in out-of-distribution test samples, and preventing potential model failures.

In the context of neural networks, uncertainty is categorized into two types: aleatoric, which is inherent in the data, and epistemic, which relates to the predictive aspects. While aleatoric uncertainty can be extracted directly from data, epistemic uncertainty is typically evaluated using methods such as Bayesian neural networks. These networks implement probabilistic priors on network weights and seek to approximate output variance. However, the Bayesian neural networks face significant challenges, including the complex inference of weight posterior distributions, the need for sampling during inference, and the difficulty in choosing appropriate weight priors.

To solve the problem mentioned above, evidential deep learning has been proposed as an alternative to Bayesian neural networks, through the learning process as one of evidence collection. In particular, evidential deep learning methodology enables the direct application of priors to the likelihood function, rather than to network weights. Training the network to produce hyperparameters of an evidential distribution effectively captures both epistemic and aleatoric uncertainties, eliminating the requirement for sampling.

In particular, the previous method ~\cite{amini2020deep} models target variables as single Gaussian distribution with unknown mean and variance and further imposes a prior of Normal-Inverse Gamma (NIG) distribution on the unknowns. An evidence prediction network is trained to predict the hyperparameters of the prior by maximizing the marginal likelihood of the target variables. This approach directly yields target predictions along with measures of both epistemic (model) and aleatoric (data) uncertainty.

\begin{figure*}[t]
  \centering
    \includegraphics[width=1\linewidth]{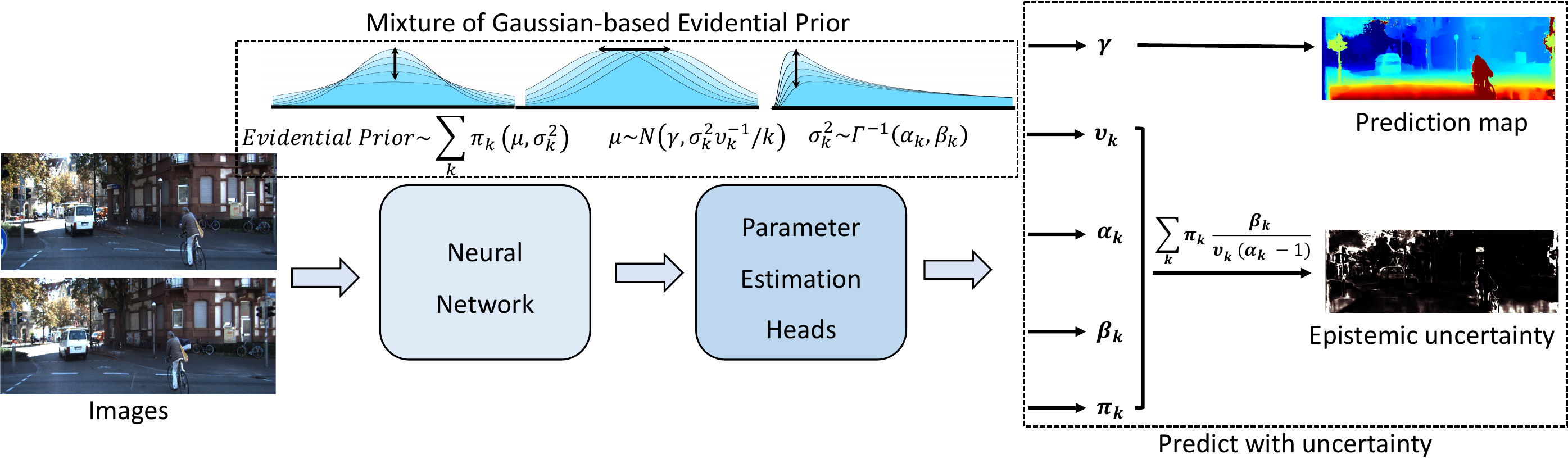}
    \vspace{-0.5cm}
    \caption{
    Illustration of the proposed pipeline for mixture-of-Gaussian based Evidential Learning. Given a set of images as input, our objective is to train a network to estimate the parameters of a mixture-of-Gaussian evidential distribution. Instead of directly predicting depth estimation maps, our approach concurrently predicts Epistemic uncertainties to estimate the evidential distributions.
    The objective is attained by hierarchically modeling the targets using a Gaussian mixture likelihood, where each component is characterized by individual likelihood parameters denoted as $(\mu, \sigma_k^2)$. Additionally, a higher-order distribution of Normal Inverse Gamma is applied over each set of the unknowns $(\mu, \sigma_k^2)$.
    }
\label{fig:architecture}
\end{figure*}

Departing from the previous models that assume a single Gaussian distribution for data in estimating uncertainties, our research introduces a more realistic hypothesis: data conforms to a mixture of Gaussian distribution to approximate more complex distribution in real life. The inspiration comes from the real application of depth estimation, where pixels of the image usually form multiple clusters. This Gaussian mixture model provides the flexibility for each image pixel to associate with multiple Gaussian distributions, allowing for the selection of the most suitable combined distributions. Utilizing the inverse-Gamma distribution as a bridge for unknown means and variances, our probabilistic framework proficiently predicts both epistemic uncertainty and aleatoric uncertainty of each component, as illustrated in Figure~\ref{fig:architecture}.

To solve the stereo matching problem, we utilize the model of STTR \cite{li2021revisiting} as our baseline, and integrate it in our mixture-of-Gaussian based deep evidential learning framework. This framework achieves improved depth estimation compared to its counterpart with a single Gaussian, effectively capturing the model uncertainty and enabling a strong cross-domain generation ability. Our method is found to achieve new state-of-the-art results on both the in-domain and cross-domain datasets.

The main contributions of this paper are summarized as follows:

\begin{enumerate}
\item We propose a mixture-of-Gaussian-based deep evidential learning framework for estimating both aleatoric and epistemic uncertainties. This framework facilitates precise regression predictions accompanied by more reliable uncertainty estimates, enhancing alignment with real-world scenarios.

\item We conduct extensive experiments in stereo matching using a variety of datasets, such as Scene Flow, KITTI 2015, and Middlebury 2014. These experiment results consistently showed enhanced accuracy compared to the baseline and demonstrated improved cross-domain generalization.

\item We perform experiments on a set of benchmark regression tasks compared to other widely used baseline methods for predictive uncertainty estimation. The results consistently demonstrate competitive prediction accuracy.

\end{enumerate}

\section{Related Works}
\subsection{Uncertainty Estimation}
In the current landscape of deep learning, particularly in safety-critical applications like autonomous vehicle navigation and medical diagnostics, accurately measuring model confidence is crucial. This need has attracted significant research into uncertainty estimation in machine learning models. Bayesian neural networks, which replace deterministic weight parameters with distributions, stand out in this field \cite{neal2012bayesian, mackay1992bayesian, kendall2017uncertainties,liu2024harmonizing,crnet,liu2021cross}. However, the extensive parameter space of these networks makes inference challenging. Approaches like variational dropout \cite{2017dropout} and Monte Carlo dropout (MC Dropout) \cite{gal2016dropout, 2016drop} have been developed to address this, with MC Dropout approximating the training process as variational inference by formulating dropout as Bernoulli distributed random variables.

Deep ensemble methods \cite{lakshminarayanan2017simple, chitta2018large, wen2020batchensemble, 2018ensemble} have also gained traction for their ability to model uncertainty by aggregating predictions from multiple models with diverse architectures. While effective, these methods can be computationally and memory-intensive, leading to the exploration of more practical solutions like pruning \cite{cavalcanti2016combining} and distillation \cite{lindqvist2020general}, or training sub-networks with shared parameters \cite{antoran2020depth}.

In contrast, deterministic neural networks offer a more efficient approach for uncertainty estimation, computing prediction uncertainties in a single forward pass \cite{mozejko2018inhibited, liu2020simple, van2020uncertainty, 2020address, Mukhoti2020Calibrating}. Notable techniques include evidential deep learning \cite{sensoy2018evidential} and prior networks \cite{malinin2018predictive}, which employ Dirichlet priors for discrete classification predictions. Deep evidential regression \cite{amini2020deep}, an extension of these methods, estimates the parameters of the normal inverse gamma distribution for regression tasks, effectively representing epistemic and aleatoric uncertainties.  Mixture of normal-inverse gamma distribution is then proposed in \cite{ma2021trustworthy} to extend the evidential learning to multimodal data.

Our research diverges from traditional models that assume a single Gaussian distribution for uncertainty estimation~\cite{amini2020deep,lou2023elfnet,gong2024learning}. We propose a more nuanced hypothesis that data conforms to a Gaussian mixture distribution, providing the flexibility for each target to associate with multiple components and enabling the selection of the most appropriate component. 
This extension of mixture distribution enables the modeling of more complex real-life data, leading to more accurate predictions and improved uncertainty estimation in complex machine learning tasks.

\subsection{Deep Stereo Matching} 
Stereo Matching is one of the complex regression tasks considered in this paper. We apply evidential learning on top of existing deep neural networks to improve the depth prediction accuracy as well as the model's cross-domain generalization.

In the evolving field of deep learning-based stereo matching, methods generally fall into two categories: CNN-based cost-volume matching and transformer-based matching. The DispNetC \cite{sceneflow} established the first fully trainable stereo matching framework using CNNs, creating a 3D cost volume through a dot product of left and right feature maps. Although efficient, this correlation operation often lacks the depth of knowledge required for optimal results. Building on GC-Net \cite{kendall2017end}, a series of studies \cite{Chang18,Zhang2019GANet,LEAStereo2020} have employed 3D hourglass convolutions to process a 4D cost volume, formed by concatenating features from stereo pairs. While effective, this approach is memory and computationally intensive.
GwcNet \cite{gwc2019} addressed these challenges by introducing group-wise correlation, creating a more compact cost volume, and achieving a better efficiency balance. ACVNet \cite{acv2022} further improved this with an attention-aware cost volume, reducing redundant information and easing the cost aggregation process.

To reduce computational demands and leverage robust semantic information, multi-scale cost volumes have been integrated into stereo matching. HSMNet \cite{yang2019hierarchical} introduced a hierarchy of volumes for processing high-resolution images. AANet \cite{Xu20}, a lighter framework, utilized a feature pyramid with multi-scale correlation volume interaction. CFNet \cite{Shen_2021_CVPR} enhanced this by constructing a cascade pyramid cost volume, narrowing the disparity search range, and refining disparity maps in a progressive manner. Recently, PCWNet \cite{shen2022pcw}, combines multi-scale 4D volumes with a volume fusion module and employs a multi-level loss for faster model convergence.

With the rise of attention mechanisms \cite{vaswani2017attention}, transformer-based deep stereo matching methods have gained prominence. STTR \cite{li2021revisiting} adopts a sequence-to-sequence approach, enforcing a uniqueness constraint and avoiding fixed disparity cost volumes. CSTR \cite{guo2022context} integrates the Context Enhanced Path module for better global information integration, addressing challenging areas such as texture-less or specular surfaces. However, despite their capacity to model long-range global information, transformer-based methods still face challenges in cross-domain validation and out-of-distribution object estimation. By incorporating deep uncertainty estimation, we aim to enhance model robustness in stereo matching tasks.
\vspace{+1mm}
\section{Method}
In this section, we will introduce our proposed mixture-of-Gaussian based Evidential Learning framework to conduct the uncertainty estimation. As depicted in Figure~\ref{fig:architecture}, given a set of input images, the primary goal is to train a network capable of estimating the parameters of the network for the task with the mixture-of-Gaussian evidential distribution as prior. Specifically, the process entails using a baseline neural network for feature extraction, followed by the acquisition of predictive target values and parameters through distinct heads integrated on top of it. Instead of direct prediction, our approach simultaneously assesses the prediction uncertainty. By enabling the model to be aware of its own prediction uncertainty, we argue that the model is less prone to overfitting and more robust for generalization. 

\subsection{Preliminary Knowledge}
Dense prediction tasks strive to conduct a regression prediction. Rather than just a target point estimation, the evidential learning \cite{amini2020deep} provides a probabilistic estimation and is able to learn both the aleatoric uncertainty and the model's epistemic uncertainty by hierarchical probabilistic modeling on the target variables. Given a dataset $\mathcal{D} =\{\mathbf{x}_i,y_i\}_1^N$, where $\mathbf{x}_i$ is the $i$th input and $y_i$ is its corresponding target.  In evidential learning, each element in the observed target, $y_i$, is supposed to follow a single Gaussian distribution, with unknown mean and variance $(\mu, \sigma^2)$. 
\begin{equation}\label{likelihood}
\{y_1,\dots, {y_i},\dots,y_N \}\sim \mathcal{N} (\mu, \sigma^2),
\end{equation}
where variable pair $({\mu}, \sigma^2)$ further follows a normal-inverse-gamma (NIG) distribution:
\begin{align} 
\label{NIG_prior}
 &p({\mu}, \sigma^2) = \cal{NIG}(\gamma,\nu,\alpha, \beta) \\ \nonumber 
 & = \frac{\beta^{\alpha}{\sqrt \nu}}{\Gamma(\alpha){\sqrt {2\pi \sigma ^2}}}\left ( \frac{1}{\sigma ^2} \right )^{\alpha+1}exp\left \{ -\frac{2\beta+\nu(\gamma-\mu)^2}{2\sigma ^2} \right\}.
\end{align}
The NIG prior above can be interpreted in two steps. Firstly, the variable $\mu$ has a normal conditional distribution:

\begin{align}
p(\mu \mid \sigma^2) &= \mathcal{N}(\gamma, \sigma^2 \nu^{-1}) \\ \nonumber
&= \frac{1}{\sqrt{2\pi \sigma^2 \nu^{-1}}} \exp\left(-\frac{\nu (\mu - \gamma)^2}{2 \sigma^2}\right)
\end{align}

with mean $\gamma$ and variance $ \sigma^2\nu^{-1}$. $\sigma^2$ then follows a marginal distribution of inverse-gamma:
\begin{align}
p(\sigma^2) = \Gamma^{-1}(\alpha, \beta) = \frac{\beta^{\alpha}}{\Gamma(\alpha)}\left ( \frac{1}{\sigma ^2} \right )^{\alpha+1}exp\left \{ -\frac{\beta}{\sigma ^2} \right\},
\end{align}
where $\Gamma(\cdot)$ is the gamma function, $\gamma\in\mathbb{R}$, $\nu>0$, $\alpha>1$ and $\beta>0$ are the hyperparameters, which determine both the location (mean) and the dispersion concentration (uncertainty) of the likelihood of the target observation $y_i$.

The target prediction i.e., $\hat y_i = \mathbb{E}(\mu)$, the aleatoric uncertainty i.e., $\mathbb{E}(\sigma^2)$ and the epistemic uncertainty i.e., $\text{Var}(\mu)$ can be  obtained  by computing the following expectations given the NIG distribution in equation (\ref{NIG_prior}):
\begin{align} \label{prediction}
{\hat y}_i = \mathbb{E}(\mu) =\int \int \mu  p(\mu,\sigma^2) d\mu d\sigma^2= \gamma,
\end{align}
\begin{align} \label{aleatoric}
\mathbb{E}(\sigma^2) = \int \sigma^2  p(\sigma^2) d\sigma^2 =\frac{\beta}{\alpha-1}, 
\end{align}
 and 
\begin{align} \label{epistemic}
\text{Var}(\mu) = \int \int (\mu - \gamma)^2  p(\mu,\sigma^2) d\mu d\sigma^2=\frac{\beta}{\nu(\alpha-1)},
\end{align} 
with $\alpha>1$.

\begin{figure*}[t]
  \centering
    \includegraphics[width=0.9\linewidth]{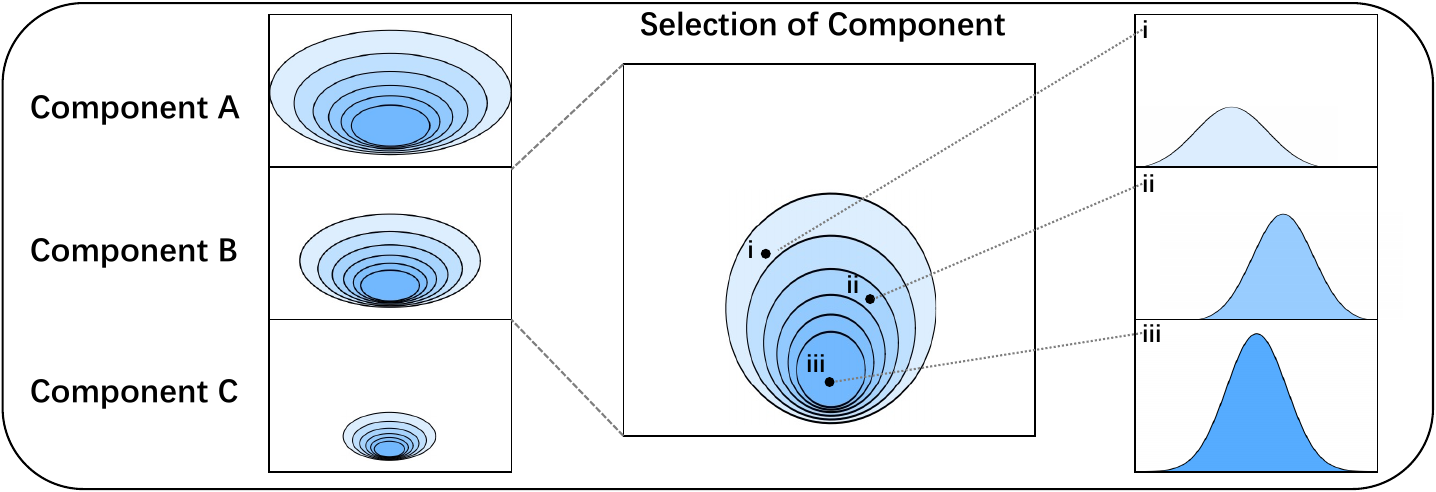}
    \caption{
    Given a dataset, our proposed mixture-of-Gaussian evidential estimation method selects the most suitable Gaussian distribution. The component distribution with a lower variance indicates concentrated data points. Conversely, datasets with diverse samples, resulting in higher variance (e.g., long-tailed distribution) can be captured by components with a higher variance. Benefiting from the mixture-of-Gaussian approach, we are able to model flexibly more complex distributed targets. The shading intensity corresponds to the probability mass, with darker shading indicating higher probability. Our objective is to train a model predicting the target $y$ from an input $x$ while incorporating an evidential prior on the likelihood to facilitate uncertainty estimation.
    }
\label{fig:distribution}
\end{figure*}

\begin{table*}[ht]
\centering
\scalebox{0.75}{\begin{tabular}{l|cccc|cccc}
\hline
 & \multicolumn{4}{c|}{RMSE} & \multicolumn{4}{c}{NLL} \\ \cline{2-9} 
 & dropout$\downarrow$ & ensembles $\downarrow$ & EL$\downarrow$ & MOG EL$\downarrow$ & dropout$\downarrow$ & ensembles $\downarrow$ & EL $\downarrow$ & MOG EL$\downarrow$ \\ \hline\hline
Boston &  \underline{2.97 ± 0.19} & 3.28 ± 1.00 &  3.06 ± 0.16 & \textbf{2.95±0.24} &2.46 ± 0.06 &2.41 ± 0.25 & \underline{2.35 ± 0.06} & \textbf{2.31±0.07}\\
Concrete &  \textbf{5.23 ± 0.12} & 6.03 ± 0.58 &  5.85 ± 0.15 & \underline{5.78±0.11} &3.04 ± 0.02 & 3.06 ± 0.18& \underline{3.01 ± 0.02} & \textbf{3.00 ± 0.00} \\
Energy &  \textbf{1.66 ± 0.04} & 2.09 ± 0.29 &  2.06 ± 0.10 & \underline{2.02±0.12} &1.99 ± 0.02 & 1.38 ± 0.22& \underline{1.39 ± 0.06} &  \textbf{1.38 ± 0.04} \\
Kin8nm  &  0.10 ± 0.00 & 0.09 ± 0.00 &  0.09 ± 0.00 & 0.10 ± 0.00 & -0.95 ± 0.01 & -1.20 ± 0.02 & \textbf{-1.24 ± 0.01} & \underline{-1.23 ± 0.01}  \\
Naval &  0.01 ± 0.00 & 0.00 ± 0.00 &  0.00 ± 0.00 & 0.00 ± 0.00 & -3.80 ± 0.01 & -5.63 ± 0.05 &  \textbf{-5.73 ± 0.07} & \underline{-5.71 ± 0.05} \\
Power &  \textbf{4.02 ± 0.04} & \underline{4.11 ± 0.17} &  4.23 ± 0.09 & 4.21 ± 0.11 &\underline{2.80 ± 0.01} & \textbf{2.79 ± 0.04}& 2.81 ± 0.07 & 2.90 ± 0.05 \\
Protein &  \textbf{4.36 ± 0.01} & 4.71 ± 0.06 &  \underline{4.64 ± 0.03} & 4.67 ± 0.00 & 2.89 ± 0.00 & 2.83 ± 0.02&  \underline{2.63 ± 0.00} & \textbf{2.61 ± 0.00}\\
Wine  &  0.62 ± 0.01 & 0.64 ± 0.04 &  \textbf{0.61 ± 0.02} & \underline{0.62 ± 0.00} & 0.93 ± 0.01 & 0.94 ± 0.12 & \textbf{0.89 ± 0.05} & \underline{0.90 ± 0.07} \\
Yacht &  \textbf{1.11 ± 0.09} & 1.58 ± 0.48 &  1.57 ± 0.56 & \underline{1.57 ± 0.54} &1.55 ± 0.03 & 1.18 ± 0.21&\underline{1.03 ± 0.19} &  \textbf{1.01 ± 0.18} \\
\hline
\end{tabular}
}

\caption{Benchmark regression tests. 
We conducted a comparative analysis between our proposed MOG evidential learning (EL) framework and traditional evidential learning (EL)~\cite{deep_evidential} approaches, as well as dropout sampling~\cite{2016drop} and model ensembling~\cite{lakshminarayanan2017simple} techniques. 
}
\label{table: edl_toy}

\end{table*}

\subsection{Mixture of Gaussian-based Evidential Learning}
To encourage a more capable and flexible distribution of targets, we further assume that
each observed target element $y_i$ follows a Gaussian mixture distribution: 
\begin{align}\label{likelihood}
 p({y_i})= \sum_{k=1}^{K}{\pi}_k\mathcal{N} (y_i \mid \mu, \sigma_k^2),
\end{align}
with ${\pi}_k$ the mixing coefficients, where
$\sum_{k=1}^{K}{\pi}_k = 1$ and 
${\pi}_k \geq 0$ for any $k$.
This simple change allows us to associate each target $y_i$ to one specific Gaussian component as shown in Figure \ref{fig:distribution}.

Following the hierarchical distributions in evidential learning, we assume the variable pair $({\mu}, \sigma_k^2)$ takes the form of the NIG distribution with different hyperparameters for different components:
\begin{align} 
\label{NIG_prior_k}
 & p({\mu}, \sigma_k^2) 
 = {\cal{NIG}}(\gamma,\nu_k,\alpha_k, \beta_k) \\ \nonumber 
 & = \frac{\beta_k^{\alpha_k}{\sqrt \nu_k}}{\Gamma(\alpha_k){\sqrt {2\pi \sigma_k ^2}}}\left ( \frac{1}{\sigma_k ^2} \right )^{\alpha_k+1} exp\left \{ -\frac{2\beta_k+\nu_k(\gamma-\mu)^2}{2\sigma_k ^2} \right\}.
\end{align}
The above prior on each component helps to model the predictive epistemic uncertainty with details given in Section \ref{uncertainty}.

\subsubsection{Evidence Maximization}
Having formalized the likelihood and the prior distributions of the target variable $y_i$, we next 
describe our approach to estimate the hyperparameters $\mathbf{m}_k=(\pi_k, \gamma,\nu_k,\alpha_k,\beta_k)$ to further calculate the prediction and the epistemic uncertainty. 

Evidence maximization, or the type II likelihood maximization of the observation targets $Y = \{y_1,\dots,y_N\}$ can be used for the hyperparameter estimation by integrating out the variable of mean and variances. The marginal distribution of one target $y_i$ can be calculated according to the likelihood in (\ref{likelihood}) and the priors in (\ref{NIG_prior_k}):
\begin{align} \label{marginal_likelihood}
& p(y_i \mid \mathbf{m}_k) \nonumber \\ 
& =\int_{\sigma_k^2}^{\infty}\int_{\mu}^{\infty}{\sum_k \pi_k}p(y_i \mid \mu,\sigma_k^2)\nonumber \\&~~~~~~~~~~~~~~~~~\times p(\mu,\sigma_k^2 \mid \gamma,\nu_k,\alpha_k,\beta_k)d\mu d\sigma_k^2 \nonumber \\  &
= {\sum_k \pi_k} \text{St}(y_i,\gamma,\frac{\beta_k(1+\nu_k)}{\nu_k\alpha_k}, 2\alpha_k).
\end{align}
Detailed derivations are given in the Appendix~\ref{sec: Appendix_A}. The above could be considered as a mixture of Student-t distribution. 

Estimations of the hyperparameters $\mathbf{m}_k=(\pi_k, \gamma,\nu_k,\alpha_k,\beta_k)$ can be found by maximizing the likelihood or minimizing the negative logarithm of the above distribution of the targets $Y$:

\begin{align}
\label{0:nllloss}
&\mathcal{L}^{\text{NLL}}(\mathbf{w}) = \ln \left\{ p(Y \mid \mathbf{m}_k) \right\}   \nonumber \\
&= \ln \left\{ \prod_{i=1}^{N} \sum_{k=1}^{K} \pi_k \text{St}(y_i, \gamma, \frac{\beta_k(1+\nu_k)}{\nu_k\alpha_k}, 2\alpha_k) \right\}   \nonumber \\
&= \sum_{i=1}^{N} \ln \left\{ \sum_{k=1}^{K} \pi_k \text{St}(y_i, \gamma, \frac{\beta_k(1+\nu_k)}{\nu_k\alpha_k}, 2\alpha_k) \right\},
\end{align}

where $1\leq i\leq N$ indicates each target $y_i$ in $Y$, $1\leq k\leq K$ indicates each component of the mixture and $K\ll N$. 

Similar to the maximization of the negative logarithm of the likelihoods of a mixture of Gaussian, there is the identifiability issue, which makes the problem non-convex and difficult to solve.

\subsubsection{Expectation and Maximization}
We resort to a similar way of expectation and maximization combined with gradient descent to find the hyperparameters. 

In expectation and maximization, we introduce a latent variable
\[z_i \sim Categorical(\boldsymbol{\pi})\] with $\boldsymbol{\pi} = [\pi_1,\dots,\pi_K]$ to indicate to which component one $y_i$ is associated and consider the problem of maximizing the likelihood for the complete data, i.e., $(Y, Z)$, with mixture coefficients $\pi_k>0$ and $\sum_k \pi_k =1$.
Suppose $p(z)=\prod_{k=1}^{K}\pi_k^{z_k}$. Each $y_i$ is then associated to one component $k$ by $z_k$:
$p(y_i) = \prod_{k=1}^{K} \left\{ \text{St}(y_i, \gamma, \frac{\beta_k(1+\nu_k)}{\nu_k\alpha_k}, 2\alpha_k) \right\}^{z_k}$.

Therefore the complete-data likelihood takes the form:
\begin{align}
\label{complete_data_likelihood}
&p(Y,Z \mid \mathbf{m}_k) \\ \nonumber
&= \prod_{i=1}^{N}\prod_{k=1}^{K}\pi_k^{z_{ik}} \left\{ \text{St}(y_i, \gamma, \frac{\beta_k(1+\nu_k)}{\nu_k\alpha_k}, 2\alpha_k) \right\}^{z_{ik}},
\end{align}
where $z_{ik}$ indicates target $i$ to component $k$.
Taking the logarithm to the above, we have 
\begin{align}
\label{complete_data_likelihood_1}
&\ln p(Y,Z \mid \mathbf{m}_k) = \\ \nonumber 
& \sum_{i=1}^{N}\sum_{k=1}^{K}z_{ik} \left \{ \ln(\pi_k) + \ln \left (\text{St}(y_i,\gamma,\frac{\beta_k(1+\nu_k)}{\nu_k\alpha_k}, 2\alpha_k) \right ) \right \}.
\end{align}
For the expectation step, the posterior probability $p(z_{ik} \mid y_i)$ that each component generates one datapoint $y_i$, i.e., $E\{z_{ik}\}$ is computed. We design the neural network of weight $
\boldsymbol w$, i.e., $f(w)$ with one special head of weight ${\boldsymbol w}_z$ to directly output the expected value of variable $z_{ik}$ given the input $\boldsymbol X$:
\begin{equation}\label{NN_z}
p_{ik}=E\{z_{ik}\}_{p(z_{ik} \mid y_i)} = {f}(\boldsymbol w, {\boldsymbol w}_z, \boldsymbol X), 
\end{equation}
for $i \in {1,\dots, N}$ and $k \in {1,\dots, K}$.

The maximization step aims to adjust the parameters of each component to maximize the probability of generating the data it is currently responsible for. After each component gets its certain amount of probability  $p_{ik}$ for each datapoint, the expectation of the complete data log-likelihood  can be evaluated:
\begin{align}
\label{expected_log_likelihood}
&E_z\{ln p(Y,Z \mid \mathbf{m}_k)\}  = \\ \nonumber
&\sum_{i=1}^{N} \sum_{k =1}^{K} p_{ik} \left \{ln(\pi_k)  + ln \left (\text{St}(y_i,\gamma,\frac{\beta_k(1+\nu_k)}{\nu_k\alpha_k}, 2\alpha_k) \right )\right \}. 
\end{align}

The inference of the parameters $\gamma,\nu_k,\alpha_k,\beta_k$, temporally ignoring the unrelated term $\pi_k$, can be found by maximizing the log-complete likelihood.  Similarly to the network construction in expectation step, we use the same baseline neural network $f(\boldsymbol{w})$ with other four heads of their own different weights to estimate $\gamma,\nu_k,\alpha_k,\beta_k$ by minimizing the loss below:
\begin{align}
\label{expected_log_likelihood1}
&\mathcal{L}^{NLL}(\mathbf{w})  \\\nonumber
&=\sum_{i}^{N} \sum_{k =1}^{K} p_{ik} \left\{ \ln \left( \text{St}
(y_i,\gamma,\frac{\beta_k(1+\nu_k)} {\nu_k\alpha_k}, 2\alpha_k) \right) \right\}
\end{align}
where $\ln \left( \text{St}(y_i,\gamma,\frac{\beta_k(1+\nu_k)}{\nu_k\alpha_k}, 2\alpha_k) \right)$ follows:
\begin{align}
\label{eq:nllloss_k}
& \mathcal{L}^{NLL_k}(\mathbf{w})  = \frac{1}{2}\log\left(\frac{\pi}{\nu_k}\right)- \alpha_k\log\left(\Omega_k\right) \\ \nonumber &~~+ \left(\alpha_k+\frac{1}{2}\right)\log\left(({y_i} -\gamma)^2\nu_k+ \Omega_k\right)+\log \Psi_k,
\end{align}
where $\Omega_k=2\beta_k(1+\nu_k)$ and  $\Psi_k=\left(\frac{\Gamma_k(\alpha_k)}{\Gamma_k\left(\alpha_k+\frac{1}{2}\right)}\right)$. It is noted that the loss corresponding to one component in (\ref{eq:nllloss_k}) is the same as that in \cite{deep_evidential}, which indicates our algorithm can be regarded as an extension of deep evidential learning in \cite{deep_evidential} to multiple components.

In addition, an empirical incorrect evidence penalty $\mathcal{L}^R$ is constructed and added to the above marginal likelihood loss in \cite{deep_evidential} to reduce evidence imposed on  incorrect predictions:
\begin{equation}
\label{eq:viewloss}
\mathcal{L}(\mathbf{w})_{u}=\mathcal{L}^{NLL}(\mathbf{w})+\lambda\mathcal{L}^R(\mathbf{w}),
\end{equation}
where the coefficient $\lambda$ is a hyper-parameter to balance these two loss terms.
The penalty follows:
\begin{equation}
\label{eq:loss_regularize}
\mathcal{L}^{R}(\mathbf{w}) = \sum p_{ik} \mid {y_i}-\gamma  \mid  \cdot (2\nu_k+\alpha_k)
\end{equation}
with $2\nu_k+\alpha_k$ being the total evidence of the prior of one component $k$ in terms of ``virtual-observations" in the conjugate prior interpretation\cite{deep_evidential}.

Finally, the hyperparameter $\pi_k$ can be simply estimated by:
\begin{align}
\label{estimate_pik}
\hat \pi_k  = 
\frac{1}{N}\sum_{i}^{N} p_{ik}, 
\end{align}
which are estimates of the mixing coefficients.

In summary, we design a base neural network to extract features from the input image and further impose five different heads on top of it with different weights to output the hyperparameters of the evidential distribution. Similar to \cite{deep_evidential}, We enforce the constraints on \(\nu_k, \beta_k\) with a softplus activation and further add 1  to $\alpha_k$ to guarantee that $\alpha_k>1$. Linear activation is used for the prediction of $\gamma$. As the network is trained in epochs, we observe the empirical convergence of our algorithm. In each step, we fit the posterior probability $p(z_{ik} \mid y_i)$ for datapoint $y_i$ in the expectation step with current neural network weights and under which we fit other hyperparameters in the maximization step by minimizing (\ref{eq:viewloss}). Finally, the weights of the neural network $f$ are updated by gradient descent.

\begin{algorithm}
\caption{Mixture-of-Gaussian based Deep Evidential Learning}\label{alg:cap}
\begin{algorithmic}
\State For a specific given task and its dataset $\mathcal{D}$, choose the backbone neural network $f(w)$ for target prediction.
\While{not converged}
    \State \textbf{Expectation step:}
    \State \quad Calculate the posterior probability $p_{ik}$ according to Eq.~(\ref{NN_z}) using the head $\mathbf{w}_z$ of $f(w)$.
    \State \textbf{Maximization step:}
    \State \quad Calculate $\gamma, \nu_k, \alpha_k, \beta_k$ using $f(w)$ with another four heads, each for one parameter.
    \State \quad Minimize the loss function in Eq.~(\ref{eq:viewloss}) by gradient descent to update the weights of $f$.
\EndWhile
\State Compute targets prediction according to Eq.~(\ref{prediction});
\State Compute data uncertainty for each component according to Eq.~(\ref{aleatoric_component});
\State Compute mixture coefficients estimation according to Eq.~(\ref{estimate_pik});
\State Compute the model uncertainty according to Eq.~(\ref{epistemic}).
\end{algorithmic}
\end{algorithm}

\subsubsection{Aleatoric and Epistemic Uncertainty}\label{uncertainty}
With the hierarchical probabilistic distribution, we are now ready to derive the prediction,  aleatoric and epistemic uncertainty. 
The target prediction i.e., $\hat y_i = \mathbb{E}(\mu)$ can be  obtained  by computing the following expectations given $\pi_k$ and the NIG distribution:
\begin{align} \label{prediction}
{\hat y}_i = \mathbb{E}(\mu) =\int \int \mu \sum_k \pi_k p(\mu,\sigma^2_k) d\mu d\sigma^2_k= \gamma.
\end{align}
The aleatoric uncertainty of each component i.e., $\mathbb{E}(\sigma^2_k)$ can be calculated by
\begin{align} \label{aleatoric_component}
\mathbb{E}(\sigma^2_k) = \int \sigma^2_k  p(\sigma^2_k) d\sigma^2_k =\frac{\beta_k}{\alpha_k-1}, 
\end{align}
and the epistemic uncertainty i.e., $\text{Var}(\mu)$ follows
\begin{align} \label{epistemic}
\text{Var}(\mu) &= \int \int \sum_k \pi_k (\mu - \gamma)^2  p(\mu,\sigma^2_k) d\mu d\sigma^2_k \nonumber \\  &=\sum_k \pi_k \frac{\beta_k}{\nu_k(\alpha_k-1)},
\end{align} 
with $\alpha_k>1$ for all $k$. Detailed derivations can be found in the Appendix~\ref{sec: Appendix_A}.

\subsubsection{Pseudo Algorithm}
The pseudo algorithm of the proposed method is given in Algorithm \ref{alg:cap}. In the experimental section, we employ a simple neural network similar to the one used in \cite{deep_evidential} for general regression tasks. Additionally, we extend its capabilities by incorporating additional predictions for $p_{ik}$. For stereo matching, we opt for a transformer-based network, i.e., STTR, and augment it with five different heads to distinguish the estimation of various hyperparameters.
\section{Experiments}
In this section, we assess the effectiveness of our Mixture of Gaussian-based Evidential Learning (MOG EL) framework in two distinct tasks: stereo matching and general regression tasks using dataset \cite{deep_evidential}. The stereo matching experiments were conducted on several datasets, including Scene Flow \cite{sceneflow}, $\&$ KITTI 2015 \cite{kitti2015}, and Middlebury 2014 \cite{scharstein2014high}. 
\subsection{General Regression Task Tests}
Following \cite{deep_evidential}, we conduct tests of the proposed method and compare it to baseline methods for predictive uncertainty estimation on datasets used in \cite{deep_evidential}, where methods of the Evidential Learning (EL) in \cite{deep_evidential}, model ensembles \cite{2018ensemble} and dropout sampling \cite{gal2016dropout} are compared based on root mean squared error (RMSE) and negative log-likelihood (NLL).

Table~\ref{table: edl_toy} presents qualitative results of RMSE and the NLL of different methods to illustrate the performance of the proposed MOG EL. The results show that our proposed method can achieve either the best indicated in bold or the second best indicated by underlining both the RMSE and the NLL metrics for most datasets.
\begin{table*}[ht]
\centering

\scalebox{0.9}{\begin{tabular}{l|cc|cc}
\hline
 & \multicolumn{2}{c|}{Disparity \textless 192} & \multicolumn{2}{c}{All Pixels} \\ \cline{2-5} 
 & EPE(px)$\downarrow$ & D1-1px(\%)$\downarrow$ & EPE(px)$\downarrow$ & D1-1px(\%)$\downarrow$ \\ \hline\hline
PSMNet~\cite{Chang18} &  0.95 & 2.71 &  1.25 & 3.25  \\
GwcNet~\cite{gwc2019} &  0.76 & 3.75  &  3.44 & 4.65  \\
CFNet~\cite{Shen_2021_CVPR} &  0.70 & 3.69  &  1.18 & 4.26 \\
PCWNet~\cite{shen2022pcw} &  0.85 & 1.94  &  0.97 & 2.48 \\
GANet~\cite{Zhang2019GANet} &  0.48 &  4.02  &  0.97 & 4.89  \\
STTR~\cite{li2021revisiting} &  0.42 & 1.37  &  0.45 & {1.38}  \\
CSTR~\cite{guo2022context} & 0.44 & 1.41  &  0.45 & 1.39  \\ 
ELFNet~\cite{lou2023elfnet}  & {0.43} & {1.35}  &  {0.44} & 1.44 \\ \hline
Ours & \textbf{0.31} & \textbf{1.01}  &  \textbf{0.33} & \textbf{1.13} \\
\hline
\end{tabular}
}

\caption{Comparison with state-of-the-art on Scene Flow~\cite{sceneflow} dataset. Our method achieves a new state-of-art performance.}

\label{table: sota-scene-flow}
\end{table*}
\begin{figure*}[t]
  \centering
\includegraphics[width=1\linewidth]{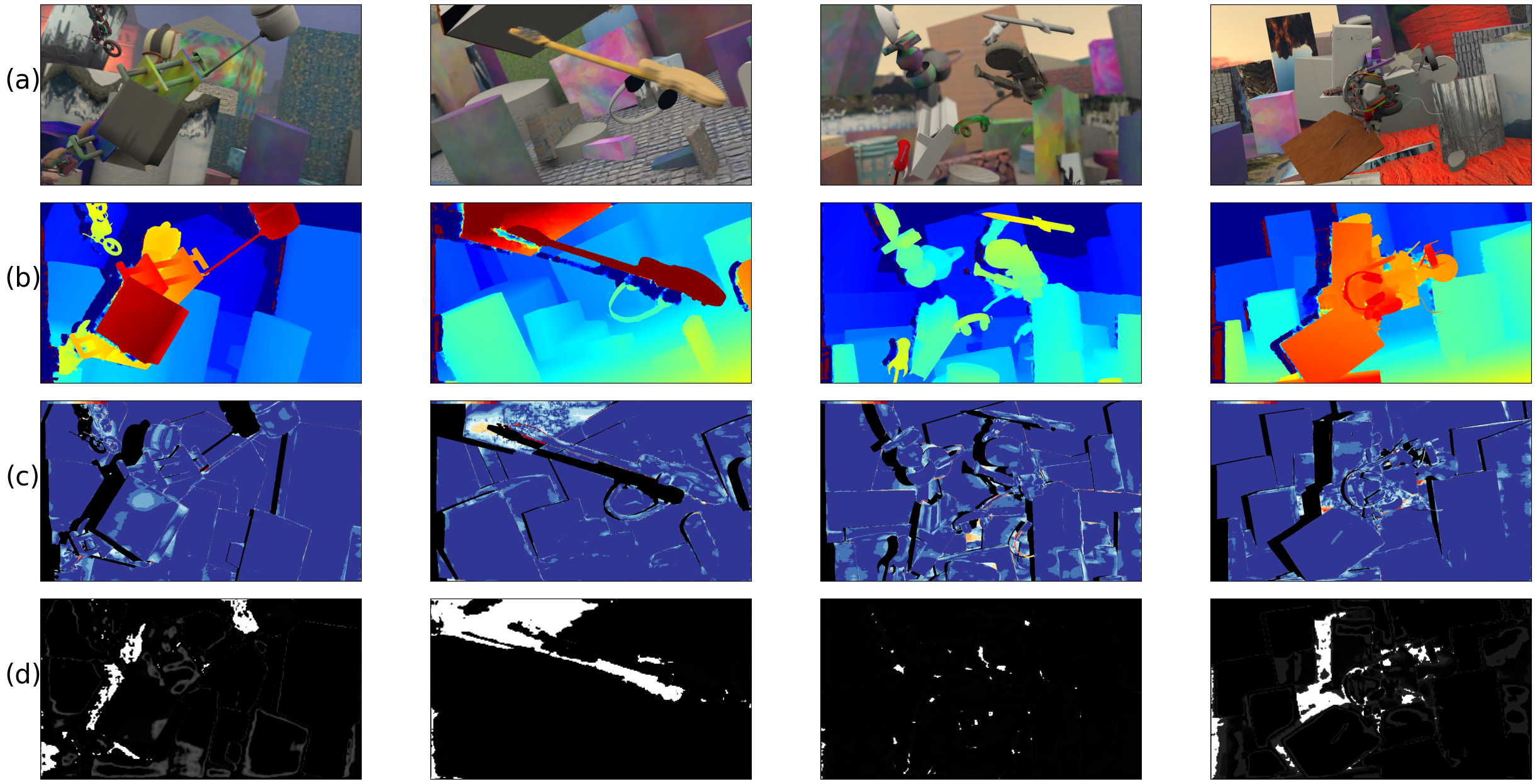}
    \caption{
    Depth Estimation and Epistemic Uncertainty. This figure includes (a) Input images; (b) Pixel-wise depth predictions; (c) Error maps; (d) Epistemic uncertainty maps.
    }
\label{fig:error_uncertainty}
\end{figure*}
\begin{figure*}[ht]
  \centering
    \includegraphics[width=1\linewidth]{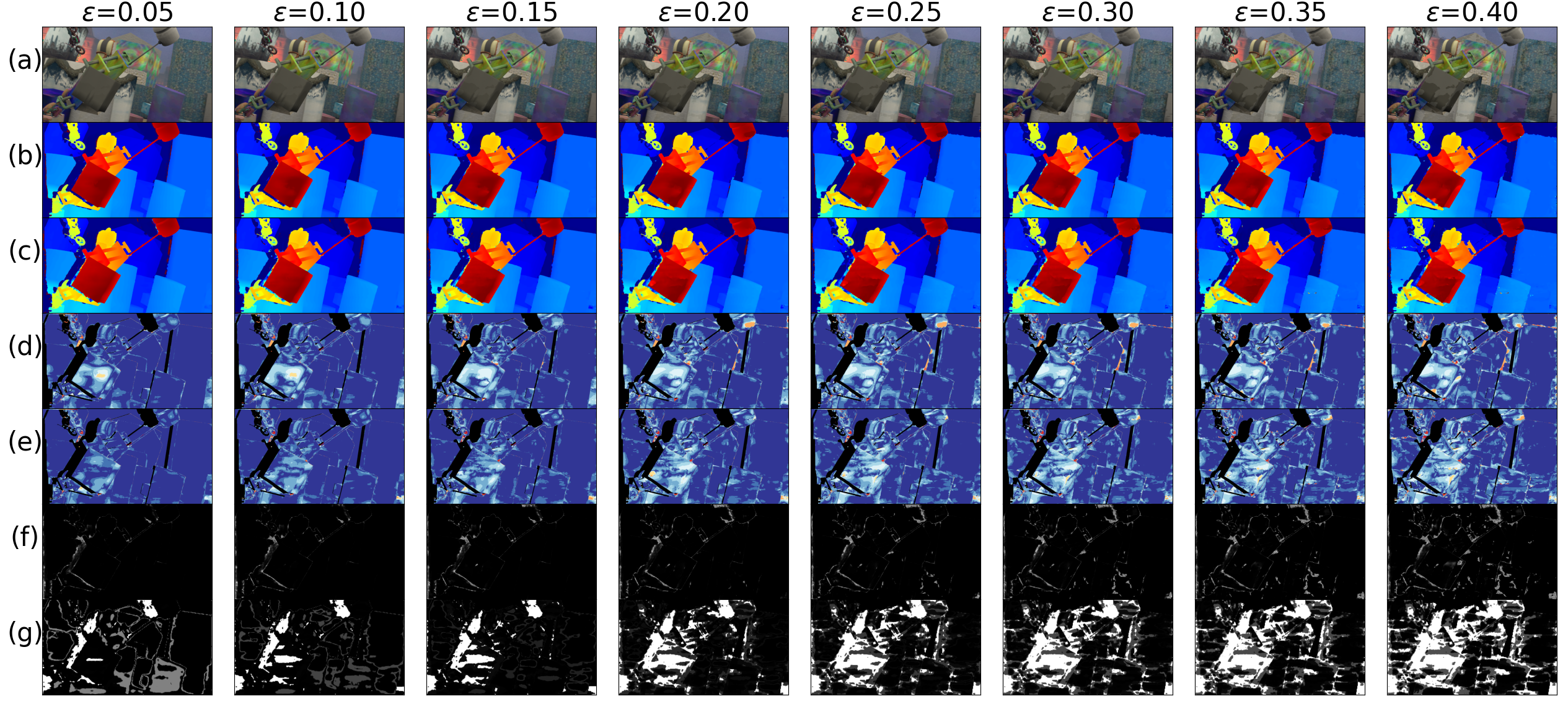}
    \caption{Effect of Adversarial Perturbation on Predictions and Uncertainty over MOG EL and EL frameworks. (a) the left input images, (b,c) predictions of EL and MOG EL, (d,e) error maps of  EL and MOG EL, and (f,g) uncertainties of  EL and MOG EL.
    }
\label{fig:uncertainty_any}
\end{figure*}
\begin{table*}[ht]
\centering
\scalebox{0.9}{
\begin{tabular}{l|cc|cc}
\hline
  & \multicolumn{2}{c|}{Middlebury 2014} & \multicolumn{2}{c}{KITTI 2015}  \\ \cline{2-5} 
& EPE(px)$\downarrow$ & 3px Err(\%)$\downarrow$ & EPE(px)$\downarrow$ & D1-3px(\%)$\downarrow$ \\ \hline\hline
PSMNet~\cite{Chang18}   & 3.05 & 13.0 &  6.59 & 16.3 \\
GwcNet~\cite{gwc2019}   & 1.89 & 8.95 & 2.21 & 12.2 \\
CFNet~\cite{Shen_2021_CVPR} & {1.69} & 7.73 &  2.27 & {5.76} \\
PCWNet~\cite{shen2022pcw}   & 2.17 & 9.09  & 1.88 & 6.03 \\
ELFNet~\cite{lou2023elfnet}  & {1.79} & {5.72} &  {1.57}  & \textbf{5.82} \\ 
STTR~\cite{li2021revisiting}  & 2.33 & {6.19} & {1.50} & 6.40 \\ \hline
Ours  &  \textbf{1.30}	 &   \textbf{5.49}  &   \textbf{1.35}	 &  {6.52} \\
 
\hline
\end{tabular}}
\caption{Cross-domain evaluation without \textit{fine-tuning} on Middleburry 2014~\cite{scharstein2014high} and KITTI 2015~\cite{kitti2015}.}
\label{table: cross_domain}
\end{table*}

\begin{table}[t]
\begin{tabular}{l|l|l|l}
\hline
Method & FLOPs(G) & Params(M) & Test Time(s) \\ \hline
Baseline   & 798.23   & 25.13     & 2.24         \\ \hline
+ MOG  &  806.79 \tiny{+8.56}     & 25.56 \tiny{+0.43}      & 2.25 \tiny{+0.01}       \\ \hline
\end{tabular}
\caption{Comparison of FLOPS, Params, and Test Time between our method and the baseline.}
\label{table: flops}
\end{table}

\subsection{Stereo Matching}
\subsubsection{Datasets and Evaluation Metrics}
\textbf{Scene Flow FlyingThings 3D subset} \cite{sceneflow}, is a synthetic dataset, that provides around 25,000 stereo frames (960$\times$540 resolution) with detailed sub-pixel ground truth disparity maps and occlusion regions. 

\textbf{KITTI 2015} datasets \cite{kitti2015}, sourced from real-life driving scenarios, include 200 training image pairs, with corresponding testing pairs and sparse disparity maps. 

\textbf{Middlebury 2014} dataset \cite{scharstein2014high} consists of high-resolution indoor stereo pairs, from which we selected the quarter-resolution images for our experiments.

\textbf{Evaluation Metrics.} 
Our evaluation metrics for disparity prediction include end-point error (EPE), the percentage of disparity outliers by 1px or 3px (D1-1px / D1-3px), and the percentage of errors exceeding 3px (3px Err).

\subsubsection{Compare with previous methods}
To validate the performance of our proposed mixture-of-Gaussian-based evidential learning method, we conducted a comparative analysis against several sota approaches in the stereo matching task. These include PSMNet \cite{Chang18}, GwcNet \cite{gwc2019}, CFNet \cite{Shen_2021_CVPR}, PCWNet \cite{shen2022pcw}, GANet \cite{Zhang2019GANet}, STTR \cite{li2021revisiting}, CSTR \cite{guo2022context}, and ELFNet \cite{lou2023elfnet}. The comparative results, as detailed in Table~\ref{table: sota-scene-flow}, the experiments are conducted on the Scene Flow dataset \cite{sceneflow}.

Our method demonstrates superior performance over these sota models in all metrics such as end-point error (EPE) and disparity outliers (D1-1px), particularly in scenarios encompassing all disparity pixels. Notably, our approach shows a significant improvement of \textbf{25\%} in EPE (0.33 vs. 0.44) and \textbf{21.5\%} in D1-1px (1.13 vs. 1.44) compared to the previously best-performing method, ELFNet \cite{lou2023elfnet}.

\begin{figure*}[t]
  \centering
    \includegraphics[width=1\linewidth]{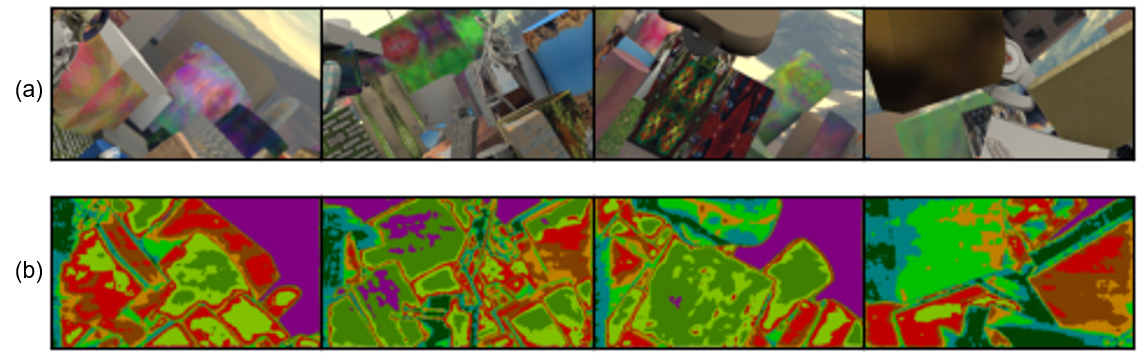}
    \caption{Illustrations of the components of the Gaussian mixture of images.
    }
\label{fig:components_illustration}
\end{figure*}

Our research underscores the enhancement in accuracy achieved by leveraging uncertainty estimation with a more realistic prior estimation network, as demonstrated by the results of both the ELFNet method \cite{lou2023elfnet} and our own. The notable performance improvements observed in our approach compared to ELFNet further validate the advantages of employing a mixture model in evidential learning. Our proposed method establishes a new state-of-the-art performance on the datasets tested.

\begin{table}[ht]
\centering
\caption{Optimal number of components of Gaussian mixture.}
\resizebox{1\linewidth}{!}{
\begin{tabular}{c|c|c|c|c|c}
\hline
 \#components& 5 & 10 & 20 & 50 & 100 \\
\hline
EPE(px) & 2.351 & 2.615 & 1.987 & 2.024 & 2.006\\
\hline
\end{tabular}
}
\label{no_Gaussian_components}

\end{table}
\subsubsection{Mixture Component Analysis}
The number of components of the Gaussian mixture is found empirically by training using a subset of 1000 images. The best component number is found to be $20$ as shown in Table. \ref{no_Gaussian_components}, where the minimal EPE error is achieved. 
Figure~\ref{fig:components_illustration} shows the components of $4$ given images, where the component $k$ for each pixel $i$ is chosen with the maximum estimated probability $p_{ik}$. These components effectively segment the objects, thereby aiding in the estimation of the depth of each object, particularly by providing clear object edges. 

\subsubsection{MOG EL against EL}
To answer the question that how the proposed mixture-of-Gaussian-based evidential learning affects the baseline, compared to the Gaussian-based evidential learning, we conduct experiments from different aspects. 

\textbf{Error maps and Uncertainty}
Qualitative results on Scene Flow using our model in are given in Figure \ref{fig:error_uncertainty}. High uncertainties are observed in occluded regions, boundary regions, and areas with large errors in the error map. This result suggests that uncertainty maps provide clues for estimating accuracy.

\textbf{Uncertainty against Adversarial perturbation}
To validate the robustness of the proposed Mixture-of-Gaussian (MOG) evidential learning framework, we conducted experiments on inputs with adversarial perturbations to challenge prediction accuracy. These perturbations were implemented on our test set using the Fast Gradient Sign Method (FGSM)\cite{goodfellow2014explaining}, with varying noise intensities ($\epsilon$). The impact of these adversarial perturbations on evidential uncertainty is illustrated in Figure \ref{fig:uncertainty_any}, which displays the predicted depth, error, and estimated pixel-wise uncertainty.

The results showcase that the STTR model with MOG EL captures increased predictive uncertainty on samples that have been adversarily perturbed. Although providing accurate depth prediction, the baseline ELFNet~\cite{lou2023elfnet} with conventional EL does not capture the uncertainty effectively. It is also shown that our model outperforms baseline in terms of more accurate depth estimations and smaller error maps.
\subsubsection{Cross-domain Generalization}

To validate the impact of our proposed mixture-of-Gaussian-based uncertainty learning on model robustness, we conducted cross-domain generalization experiments. Specifically, we evaluated the pre-trained model, originally trained on the synthetic Scene Flow dataset \cite{sceneflow}, on real-world datasets (Middlebury 2014, and KITTI 2015) in a zero-shot setting. The results of these experiments are detailed in Table~\ref{table: cross_domain}.

The results indicate that the proposed mixture-of-Gaussian-based uncertainty learning exhibits superior generalization capabilities compared to existing state-of-the-art models on real-world datasets. Notably, when benchmarked against the STTR baseline \cite{li2021revisiting}, our model achieved a significant improvement in the end-point error (EPE) score (1.3 vs. 2.33, a \textbf{44.2\% improvement}) and in the 3px error score (5.49 vs. 6.19, a 11.3\% improvement) on the Middlebury 2014 dataset \cite{scharstein2014high}. This demonstrates the effectiveness of our approach in enhancing the robustness and generalization ability of deep learning models in cross-domain applications.

\subsubsection{The comparison of computation}  
As illustrated in Table~\ref{table: flops}, we provide a comparative analysis of inference time and memory usage. Our proposed method introduces only a marginal increase in computation compared to the baseline (ELFNet~\cite{lou2023elfnet}). 
\vspace{-2mm}
\section{Conclusion}
In this paper, we introduce a novel mixture-of-Gaussian-based deep evidential learning framework for deep neural networks to effectively address the challenges in accurately estimating model uncertainties. By moving away from traditional single Gaussian distribution assumptions and instead modeling data as conforming to a Gaussian mixture distribution, the framework allows for more accurate and flexible pixel-level predictions. The use of the inverse-Gamma distribution as an intermediary enhances the model's proficiency in predicting both aleatoric and epistemic uncertainties. Extensive experiments in stereo-matching and general regression tasks across multiple datasets demonstrated that our method outperforms existing models, especially in the stereo-matching task where new state-of-the-art results are achieved for both in-domain and cross-domain depth estimation.

\section*{} \textbf{Data Availability Statement.} 
We conducted experiments using three datasets: Scene Flow \cite{sceneflow}, KITTI 2015 \cite{kitti2015}, and Middlebury 2014 \cite{scharstein2014high}. The Scene Flow dataset is accessible at \url{https://lmb.informatik.uni-freiburg.de/resources/datasets/SceneFlowDatasets.en.html}, the KITTI 2015 dataset at \url{https://www.cvlibs.net/datasets/kitti/eval_scene_flow.php?benchmark=stereo}, and the Middlebury 2014 dataset at \url{https://vision.middlebury.edu/stereo/data/scenes2014/}. These sources provide the necessary data for our analysis and are available for public access.
\section*{Appendix} \label{sec: Appendix_A}
As $\mu$ follows conditional normal distribution ${\mathcal{N} (\gamma, \sigma^2_k\nu^{-1}_k)}$ given the variance $\sigma^2_k$ with a probability $\pi_k$, the prediction ${\hat y}_i  = \mathbb{E}(\mu)$ can be computed by the expectation of $\mu$ from the joint distribution:
\begin{align} 
{\hat y}_i & = \mathbb{E}(\mu) =\int \int \sum_k \pi_k p(\mu,\sigma^2_k) d\mu d\sigma^2_k \\ \nonumber &= \sum_k \pi_k \int \int \mu  p(\mu \mid \sigma^2_k)p(\sigma^2_k) d\mu d\sigma^2_k \\ \nonumber &= \sum_k \pi_k \int p(\sigma^2_k) \int \mu  p(\mu \mid \sigma^2_k) d\mu d\sigma^2_k \\ \nonumber &=  \sum_k \pi_k \int p(\sigma^2_k) \left \langle \mu  \right \rangle_{\mathcal{N} (\gamma, \sigma^2_k\nu^{-1}_k)}   d\sigma^2_k \\ \nonumber &= \gamma \sum_k \pi_k \int  p(\sigma^2_k)  d\sigma^2_k \\ \nonumber & = \gamma,
\end{align}
where $\left \langle \mu  \right \rangle_{\mathcal{N} (\gamma, \sigma^2_k\nu^{-1}_k)}$ denotes the expectation of $\mu$ given its distribution ${\mathcal{N} (\gamma, \sigma^2_k\nu^{-1}_k)}$, where $\gamma$ is shared by all $k$, $\int  p(\sigma^2_k)  d\sigma^2_k = 1$ and $\sum_k \pi_k =1$.

The variance of each component can be calculated by the same way as evidential learning, which is:
\begin{align} 
&\mathbb{E}(\sigma^2_k) = \int \sigma^2_k  p(\sigma^2_k) d\sigma^2_k \\ \nonumber 
& = \int  \frac{\sigma^2_k\beta^{\alpha_k}}{\Gamma(\alpha_k)}\left ( \frac{1}{\sigma ^2_k} \right )^{\alpha_k+1}exp\left \{ -\frac{\beta_k}{\sigma_k ^2} \right\} d \sigma^2_k \\ \nonumber 
& \overset{{x := \sigma^2_k}} {=} \frac{\beta_k^{\alpha_k}}{\Gamma(\alpha_k)} \int x^{-(\alpha_k-1)-1} e^{ -\beta_k x} dx \\ \nonumber 
& = \frac{\beta_k^{\alpha_k}{\Gamma(\alpha_k-1)}}{\Gamma(\alpha_k){\beta_k^{\alpha_k-1}}} \int \frac{\beta_k^{\alpha_k-1}}{\Gamma(\alpha_k-1)} x^{-(\alpha_k-1)-1} e^{ -\beta_k x} dx \\ \nonumber
& = \frac{\beta_k^{\alpha_k}{\Gamma(\alpha_k-1)}}{\Gamma(\alpha_k){\beta_k^{\alpha_k-1}}} \\ \nonumber
&=\frac{\beta_k}{\alpha_k-1}, 
\end{align}
with $\alpha_k>1$, where $x := \sigma^2_k$ denotes to substitute $\sigma^2_k$ by $x$.

Following similar procedures for the prediction estimation, the model uncertainty  $\text{Var}(\mu)$ can be computed as follows: 
\begin{align} 
\text{Var}(\mu) & = \int \int \sum_k \pi_k (\mu - \gamma)^2  p(\mu,\sigma^2_k) d\mu d\sigma^2_k \\ \nonumber &= \int \sum_k \pi_k p(\sigma^2_k) \left \langle (\mu -\gamma)^2  \right \rangle_{\mathcal{N} (\gamma, \sigma^2_k\nu^{-1}_k)}   d\sigma^2_k \\ \nonumber  & = \int \sum_k \pi_k \left \langle \mu^2  \right \rangle_{\mathcal{N} (\gamma, \sigma_k^2\nu^{-1}_k)} p(\sigma^2_k) d\sigma^2_k  - \gamma^2 \\ \nonumber & = \int \sum_k \pi_k \bigg (\gamma^2 + \frac{\sigma^2_k}{v_k}\bigg ) p(\sigma^2_k) d\sigma^2_k   - \gamma^2 \\ \nonumber & = \sum_k \pi_k \int \frac{\sigma^2_k}{v_k} p(\sigma^2_k)d\sigma^2_k \\ \nonumber & =\sum_k \pi_k \frac{\beta_k}{\nu_k(\alpha_k-1)},
\end{align} 
with $\alpha_k>1$, where $\left \langle \mu^2  \right \rangle_{\mathcal{N} (\gamma, \sigma^2_k\nu^{-1}_k)}$ denotes the expectation of $\mu^2$ given its distribution ${\mathcal{N} (\gamma, \sigma^2_k\nu^{-1}_k)}$ and $\sum_k \pi_k =1$.
\subsection*{Marginal likelihood of Target variable}
The marginal likelihood of the target variable $y_i$ given the parameters of $\gamma,\pi_k, \nu_k,\alpha_k,\beta_k$ can be achieved by integrating out $\mu$ and $\sigma^2_k$:

\begin{align*} 
&p(y_i \mid \gamma,\pi_k, \nu_k,\alpha_k,\beta_k) \nonumber \\
&= \int \int \sum_k \pi_k p(y_i \mid \mu,\sigma^2_k)p(\mu,\sigma^2_k \mid \gamma,\nu_k,\alpha_k,\beta_k) d\mu \, d\sigma^2_k \nonumber \\
& = \sum_k \pi_k \int \int \sqrt{\frac{1}{2\pi\sigma^2_k}}\exp\left \{-\frac{(y_i-\mu)^2}{2\sigma^2_k}\right\} \nonumber \\
& ~~~~\quad \times \frac{\beta_k^{\alpha_k}\sqrt{\nu_k}}{\Gamma(\alpha_k)\sqrt{2\pi \sigma^2_k}}\left( \frac{1}{\sigma^2_k} \right)^{\alpha_k+1} \nonumber \\
& ~~~~\quad \times \exp\left \{-\frac{2\beta_k+\nu_k(\gamma-\mu)^2}{2\sigma^2_k} \right\} \, d\mu \, d\sigma^2_k \nonumber \\
&= \sum_k \pi_k \int \frac{\beta_k^{\alpha_k}\sqrt{\nu_k}\sigma_k^{-3-2\alpha_k}}{\Gamma(\alpha_k)\sqrt{2\pi(1+\nu_k)}} \nonumber \\
&~~~~\quad \times \exp\Biggl\{-\Biggl(\frac{2\beta_k+\nu_k(y_i-\gamma)^2}{1+\nu_k}\Biggr)\Biggr/(2\sigma^2_k)\Biggr\} \, d\sigma^{2}_k \nonumber \\
&= \sum_k \pi_k \frac{\Gamma\left(\frac{1}{2}+\alpha_k\right)}{\Gamma(\alpha_k)} \frac{(2\beta_k(1+\nu_k))^{\alpha_k}}{\sqrt{\nu_k/\pi}} \nonumber \\
&~~~~\quad \times (\nu_k(y_i-\gamma)^{2}+2\beta_k(1+\nu_k))^{-\left(\frac{1}{2}+\alpha_k\right)} \nonumber \\
&= \sum_k \pi_k \text{St}(y_i,\gamma,\frac{\beta_k(1+\nu_k)}{\nu_k\alpha_k}, 2\alpha_k).
\end{align*}

{\small
\bibliography{sn-bibliography}
}
\end{document}